\documentclass[10pt,letterpaper]{article}

\usepackage{cogsci}
\usepackage{xcolor}
\usepackage{graphicx}
\usepackage{amsmath}
\usepackage{amssymb}
\usepackage{dsfont}
\DeclareMathOperator*{\argmax}{arg\,max}
\usepackage{amsfonts}
\usepackage{stmaryrd}
\usepackage{arydshln}
\usepackage{hyperref}
\cogscifinalcopy 

\usepackage{pslatex}
\usepackage{apacite}
\usepackage{float} 

\title{Extending rational models of communication from beliefs to actions}

\author{{\large \bf Theodore R. Sumers\textsuperscript{1}, Robert D. Hawkins\textsuperscript{2}, Mark K. Ho\textsuperscript{1}, Thomas L. Griffiths\textsuperscript{1,2}} \\
   \textsuperscript{1}Department of Computer Science, Princeton University \\
      \textsuperscript{2}Department of Psychology, Princeton University \\
      \texttt{\{sumers, rdhawkins, mho, tomg\}@princeton.edu}
 
}

\begin{document}

\maketitle

\begin{abstract}
Speakers communicate to influence their partner's \emph{beliefs} and shape their \emph{actions}.
Belief- and action-based objectives have been explored independently in recent computational models, but it has been challenging to explicitly compare or integrate them.
Indeed, we find that they are conflated in standard referential communication tasks.
To distinguish these accounts, we introduce a new paradigm called \textit{signaling bandits}, generalizing classic Lewis signaling games to a multi-armed bandit setting where all targets in the context have some relative value.
We develop three speaker models: a \textit{belief}-oriented speaker with a purely informative objective; an \textit{action}-oriented speaker with an instrumental objective; and a \textit{combined} speaker which integrates the two by inducing listener beliefs that generally lead to desirable actions.
We then present a series of simulations demonstrating that grounding production choices in future listener actions results in relevance effects and flexible uses of nonliteral language.
More broadly, our findings suggest that language games based on richer decision problems are a promising avenue for insight into rational communication.

\end{abstract}

\section{Introduction}
\label{introduction_section}
\begin{quote}
``Language is used for doing things.''~\cite[p. 3]{clark1996using}
\end{quote}
But \textit{what} things? 
Broadly, accounts of communicative goals have been formulated in terms of listener \emph{beliefs} and \emph{actions}: ``Alan is speaking with the aim of getting Barbara to understand him and to act on that understanding''~\cite[p. 11]{clark1996using}. But how do these aims relate? Is language primarily a tool for informing others, shaping their actions, or some combination of the two?

Classical accounts have emphasized \textit{beliefs}, framing communication as information transfer between speaker and listener~\cite{grice1975logic}. This is reflected in the recent Rational Speech Acts (RSA) framework, which typically defines a speaker's utility in terms of epistemic objectives like informativeness~\cite{frank2012pragmatic, Kao2014nonliteral, yoon2016polite}. A related perspective from the connectionist literature defines meaning in terms of effects on the listener's latent representations~\cite{elman2009}. Yet beliefs themselves are an unsatisfying objective, as they are imperceptible and have no real-world consequences. 
Grice himself suggested informativeness should be generalized to ``influencing or directing the actions of others''~\cite[p. 47]{grice1975logic}. 

Under an alternative \textit{action}-oriented view, communication is an extension of an agent's basic capability to interface with the world, allowing a speaker to act indirectly through others. This has been explored in game-theoretic pragmatics~\cite{qing2015variations,benz2007, franke2016_probabilisticpragmatics} and emergent communication~\cite{lazaridou2020emergent}.  Practical natural language interfaces, including instruction-following~\cite{Tellex2011} and task-oriented dialogue systems~\cite{chen2017dialogue}, typically learn direct mappings between commands and agent actions. While such imperative language is effective at inducing immediate actions, it cannot offer a full account of communication. Humans clearly employ more sophisticated strategies to ``program'' others~\cite{lupyan2016_language_programs}.

To reconcile these perspectives, we propose a unified computational model that integrates both action-based and belief-based objectives: communication \textit{operates} by influencing intermediate beliefs, but its \textit{objective} is to shape downstream actions in the world. Under this \emph{combined} model, speakers reason about the decision problem a listener faces, how utterances may change their latent beliefs, and finally how those beliefs give rise to actions. They then choose utterances to induce belief states that maximize value in expectation over possible actions. Varying the scope of the decision problem allows reasoning at different time horizons, from ``nudges''~\cite{thaler2009nudge} that encourage specific actions (e.g. getting someone to close a window by saying ``It's chilly in here'') to intervening on norms more broadly~\cite{tomasello2016cultural}. 

We first extend the Rational Speech Acts (RSA) framework \cite{frank2012pragmatic} to account for action-based objectives. We then introduce a new task paradigm, \textit{signaling bandits}, which combines traditional signaling games~\cite{lewis1969convention} with multi-armed bandits~\cite{sutton2018reinforcement}. 
In Simulation 1, we show that standard reference games are a special case of the signaling bandit framework and conflate speaker objectives. 
In Simulation 2, we find that action-based speakers provide decision-relevant information, but employ hyperbolic and false utterances that improve decisions at the cost of distorting listener beliefs. 
Finally, Simulation 3 shows that when the Combined speaker reasons about a larger set of possible actions, it becomes more truthful and produces utterances that improve decision-making over the full distribution of contexts the listener could face. These results suggest that integrated reasoning over listener belief states and actions creates a pressure to transmit appropriately generalizable information.

\section{Action-Grounded Speaker Models}

Suppose a friend is about to step into a busy street. We have a brief moment to say something-- but what should we say?
The basic principle of \emph{informativeness} suggests that we aim to reduce their uncertainty about the world as much as possible. 
Without a notion of \emph{relevance} \cite{sperber1986relevance}, however, it would be equally informative to mention a passing cloud or an incoming car.
Recent accounts have formalized relevance as a \emph{question under discussion} (QUD) that collapses the utility of the listener's beliefs to a coarse-grained state, such as the current state of the road ~\cite<e.g.>{Kao2014nonliteral,hawkins2015you}.
But our friend may benefit from a more outcome-grounded notion of relevance, projecting not to epistemic states but to the \emph{decision problem} they face \cite{benz2007}: intuitively, regardless of what they know, we want them to cross safely.

A simple way to ground relevance in the decision problem is to maximize the likelihood that the listener will take a specific action, effectively using that action as the QUD. 
This model, which we call the \emph{action}-oriented speaker, would suggest something like ``Cross after that car passes!''
However, rather than choosing a premeditated action from our own perspective, we could instead give the listener information that allows them to act optimally from \emph{their} perspective. 
This \textit{combined} speaker aims to maximize the full expected utility of the listener's actions under the induced beliefs. Such a speaker might say ``Look both ways!'' which encourages our friend to check for cars outside of our own field of view and (incidentally) generalizes across intersections.

\noindent\textbf{Belief-oriented speaker}. 
We formalize these models by first introducing the Rational Speech Acts framework~\cite{frank2012pragmatic}, which instantiates Gricean maxims \cite{grice1975logic} as recursive social inference. 
In this framework, speakers have knowledge of the world state $w$ and choose between utterances $u$ proportionally to their utility $U(u,w)$, where $\beta_S$ is a soft-max parameter controlling speaker optimality:
\begin{equation}
    P_S(u \mid w) \propto \text{exp}\{\beta_S \cdot U(u, w)\}
\end{equation}
Typically, this utility is defined in terms of the listener's \emph{beliefs}, their information gain about the state $w$: 
\begin{equation}
    U_\text{Belief}(u \mid w) = \log P_L(w \mid u)
\label{eq_belief_speaker}
\end{equation}
This utility requires the speaker to reason about the listener's expected beliefs after hearing the utterance: 
\begin{equation}
    P_L(w \mid u) \propto \delta_{\llbracket u \rrbracket (w)} P(w)
\label{eq_intro_listener_belief_update}
\end{equation} 
where $\delta_{\llbracket u \rrbracket(w)}$ represents the meaning of $u$, evaluating to one when utterance $u$ is true of $w$ and zero otherwise. 
This formulation optimizes for accurate listener beliefs, but lacks a notion of relevance. 

\noindent\textbf{Action-oriented speaker}. 
Our second model extends the basic RSA framework to reason about actions a listener could take in the environment. Rather than adding an additional objective to the epistemic utility, we \textit{ground} this objective in a decision problem.
Specifically, we re-formulate the listener as a reinforcement learning (RL) agent~\cite{sutton2018reinforcement} with a set of possible \textit{actions} that may be taken, $\cal{A}$. At each point in time, a subset of those actions are available to the listener, which we call an \textit{action context} $A \subseteq \cal{A}$.
We assume the listener will choose actions to maximize a \textit{reward} function, where the scalar reward value associated with each action is defined by the world state: $R: \mathcal{A} \times W \rightarrow \mathbb{R}$. 
We thus define the listener's policy $\pi_L$ in an action context $A$ to be
\begin{equation}
    \pi_L(a \mid u, A) \propto \exp\{\beta_L \cdot R_L(a, u)\}
\label{eq_listener_action_rule}
\end{equation}
where $\beta_L$ is the softmax optimality and $R_L$ is the listener's expected reward for action $a\in A$ after hearing utterance $u$.
We define expected reward with respect to their posterior beliefs about the likely state of the world:
\begin{equation}
    R_L(a, u) = \sum_{w \in W}  R(a, w) P_L(w \mid u)
\label{eq_listener_inferred_rewards}
\end{equation}

In this work, we assume speakers are cooperative and have access to a ground-truth world state $w$. 
Action-based rational speakers reason about how their utterances will affect listener actions (Eqs.~\ref{eq_intro_listener_belief_update},~\ref{eq_listener_action_rule},~\ref{eq_listener_inferred_rewards}) and communicate to maximize the listener's reward. 
We first describe a ``pure'' \emph{action}-oriented speaker, which chooses the highest-reward action $a^* \in A$ and optimizes the probability of the listener taking that action:
\begin{equation}
    a^* \triangleq \argmax_{a\in A} R(a, w)
\end{equation}
\begin{equation}
    U_\text{Action}(u \mid A, a^*) = \log[\pi_L(a^* \mid u, A)]
\label{eq_action_speaker}
\end{equation}
Critically, rather than aiming for high-performing beliefs in general, this speaker only considers the listener's beliefs insofar as they are relevant for producing the pre-selected action. Intuitively, this can be thought of as imperative language. 

\noindent\textbf{Combined speaker}. 
Our final speaker unifies belief- and action-oriented objectives by optimizing over both, inducing beliefs which are likely to maximize rewards \emph{in general}:
\begin{equation}
    U_\text{Combined}(u \mid A, w) = \sum_{a \in A} \pi_L(a \mid u, A) R(a, w)
\label{eq_combined_speaker}
\end{equation}
Effectively, this utility shifts the locus of decision-making to the listener: the combined speaker treats them as an independent agent and optimizes their beliefs, rather than choosing an action for them.\footnote{Note our two action-oriented speakers can be recovered from a more general utility introducing an additional soft-max over the reward term $R(a,w)$, yielding the Action utility with $\beta\rightarrow \infty$ and the Combined utility with $\beta=1$.} We now introduce an experimental setting to explore these speaker models.

\section{Signaling Bandits}
We define a new language game which enables study of the speaker models described above.
In this setting, speakers cannot directly signal a unique correct action because all actions have some relative value. 
They must instead supply partial information to guide decision making.
We first review the structure of Lewis signaling games and note their limitations. We then describe multi-armed bandits, a setting studied in reinforcement learning. Finally, we combine the two to produce a new game, \textit{signaling bandits}. 

\subsection{Lewis Signaling Games} 
Lewis signaling games are two-player collaborative settings with a speaker and a listener~\cite{lewis1969convention}. Following the notation introduced previously, a signaling game is defined by a world state $w$, action context available to the listener, $A$, and utterances available to the speaker, $\cal{U}$. There is one action $a^* \in A$ with a positive reward; other actions have zero reward. The world state $w$ implies the correct action $a^*$. The speaker knows $w$ but the listener does not. During gameplay, the speaker chooses an utterance $u\in \cal{U}$ and sends it to the listener. The listener updates their beliefs, $P_L(w \mid u)$, and uses the posterior to choose an action, $\pi_L(a \mid u, A)$.

Signaling games formalize the coordination problem underlying communication \cite{krahmer2012referringsurvey, frank2012pragmatic}. However, the interplay of beliefs, actions, and rewards is highly constrained. The state of the world $w$ is synonymous with the correct action $a^*$, and players are indifferent over other actions. It is thus impossible to discriminate the three speaker objectives defined above (Eq.~\ref{eq_belief_speaker},~\ref{eq_action_speaker},~\ref{eq_combined_speaker}).\footnote{But see~\citeA{qing2015variations} for evidence in favor of action-oriented speakers.} For a richer decision-making setting, we turn to multi-armed bandits.

\subsection{Multi-Armed Bandits} 
A multi-armed bandit is a single-player sequential game. In each round, the player takes an action and receives a scalar reward~\cite{sutton2018reinforcement}. Players seek to maximize their rewards, but are initially ignorant of the reward structure. Over multiple rounds, they must balance exploration (choosing a new action to learn its reward) with exploitation (choosing the most valuable known action). Because payoffs are scalar, decisions are more nuanced than Lewis games.

Contextual bandits extend this to study learning via abstract information. Actions are now characterized by features, and rewards are defined with respect to these features. Formally, a feature function $\phi$ describes actions: $\phi: A \rightarrow \mathbb{R}^K$. Rewards are then defined as a function of these features: $R: \phi(a) \rightarrow \mathbb{R}$. Thus, rather than learn about the reward of a specific action, players can learn about the reward of a \textit{feature} which applies to many actions. For example, an animal might learn that ripe yellow bananas are high-reward, while rotten brown bananas are low-reward. Associating the payoff with the color (a feature) rather than the banana (a specific action) allows knowledge to transfer to new settings (the next banana). Contextual bandits have been studied extensively in reinforcement learning and to a lesser degree for emergent communication~\cite{donaldson2007evolution}. Yet to our knowledge, they have not been used to study communication with an existing language. In the next section, we introduce a two-player version of this game. 

\subsection{Signaling Bandits}

We combine the communication of Lewis games with the reward structure of contextual bandits to create a new game, \textit{signaling bandits}. Unlike Lewis games, speakers no longer communicate concrete information (which action is correct). Instead, they communicate abstract information (how much features are worth). We now describe basic gameplay.

\begin{figure}[t!]
\centering
\includegraphics[width=7.5cm]{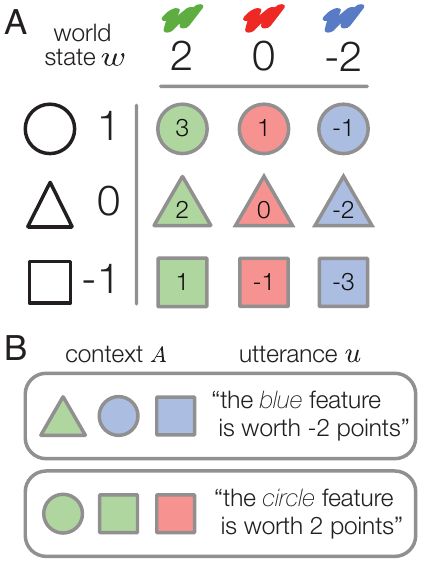}
\caption{Signaling bandits combine Lewis signaling games with multi-armed bandits. (A) The world $w$ is defined by correspondences between features and rewards (table margins), which combine additively to create possible actions $\cal{A}$ (table contents). (B) Two contexts $A$ with example utterances. The first utterance is true and useful (encouraging the listener to avoid the blue objects). The second is false (circles are worth 1). It is locally useful (encouraging the listener to choose the circle) but will lead the listener astray in other contexts.} 
\label{fig_game_features}
\end{figure}

As in Lewis games, signaling bandits are two-player games with a speaker and a listener. Each game is defined by a world state $w$, a set of all possible actions $\cal{A}$ and a set of speaker utterances $\cal{U}$. In each round, the listener faces an action context $A \subseteq \cal{A}$. However, \textit{unlike} Lewis games, there is no single ``correct'' action. Instead, as in contextual bandits, each action has a scalar reward defined by the world state $w$. Here, we assume features are indicator variables over actions: 
\begin{equation}
    \phi: A \rightarrow \{0,1\}^K
\end{equation}
Rewards $R$ are linear over these features, parameterized by $w$:
\begin{equation}
    R(a, w) = w^\top\phi(a).
\label{eq_rewards}
\end{equation}
Concretely, this means the world state $w$ is a vector encoding the reward associated with each feature. A listener with full world knowledge (every element of $w$) can calculate exact rewards using Eq.~\ref{eq_rewards} and thus select the optimal action in any context. Fig.~\ref{fig_game_features}A depicts this visually: $w$ defines the value of individual shapes and colors (table margins), which in turn determines the reward for each possible action in $\cal{A}$ (table contents). The value of an individual feature (one element of $w$) thus constitutes \textit{partial} knowledge about the world.

The speaker helps the listener by providing such partial knowledge. Formally, ${\mathcal{U}}$ is a set of tuples of the form  $\langle\mathds{1}_K, \mathbb{R}\rangle$ which specify a given feature and scalar value. As shown in Fig.~\ref{fig_game_features}, these are messages like $\langle$Blue, -2$\rangle$ or $\langle$Circle, 2$\rangle$. Speakers choose utterances and send them to the listener. The (literal) listener updates their beliefs by setting the corresponding feature to the transmitted scalar value:
\begin{equation}
    p_L(w_L \mid u) = p_L(w_L \mid u_{\mathds{1}_K} =  u_{\mathbb{R}})
\label{eq_listener_belief_update}
\end{equation}
and then chooses an action from the available set $A$ according to their posterior belief over rewards (Eqs.~\ref{eq_listener_action_rule}~and~\ref{eq_listener_inferred_rewards}).

Transmitting partial knowledge and constructing different contexts $A \subseteq \cal{A}$ introduces several important dynamics. First, it induces relevance effects, since different knowledge will be useful in different contexts. For example, in Fig.~\ref{fig_game_features}B, $\langle$Green, 2$\rangle$ would improve decision making in both contexts while $\langle$Blue, -2$\rangle$ would only be relevant for the top one. Second, it accommodates nonliteral language naturally, as false beliefs can yield good decisions. In the bottom context of Fig.~\ref{fig_game_features}B, a false message $\langle$Circle, 2$\rangle$ maximizes the probability of the listener choosing the optimal action. Finally, it allows us to explore \textit{generalization}: whether a listener's beliefs facilitate good decision making over other action contexts constructed from the same world. A bias towards communicating generalized information is implicated in cultural learning~\cite{csibra2009natural, tessler2019language}; thus, modeling the dynamics of speaker objectives and resulting generalization performance is of significant theoretical interest. 

Signaling bandits creates clean distinctions between listener beliefs $P_L(w \mid u)$, the optimal action $a^*$, and the value of individual actions, $R(a, w)$. This allows for meaningful differences between speaker models (Eqs.~\ref{eq_belief_speaker},~\ref{eq_action_speaker},~\ref{eq_combined_speaker}). In the following section, we use simulations to illustrate this. We return to extensions beyond basic gameplay in the General Discussion.

\section{Simulations}
We perform three simulations within the signaling bandits framework. We first describe the general procedure and metrics used to measure speaker behaviors. For all simulations, we set $\beta_L=3, \beta_\text{Belief}=3, \beta_\text{Action}=3$, and $\beta_{\text{Combined}}=2$. Speaker optimality does not affect the qualitative results; we analyze optimal speakers ($\beta \rightarrow \infty$) at the end of this section. We assume listeners have uniform priors over feature rewards, and allow speakers to send only a single message.

\noindent\textbf{Procedure}. For each simulation, we define a world state $w$ and set of allowable utterances $\cal{U}$. For an action context $A$, we first compute each speaker's distribution over utterances $u \in \mathcal{U}$ as defined by Eqs.~\ref{eq_belief_speaker},~\ref{eq_action_speaker}, and ~\ref{eq_combined_speaker}. We then compute the listener's resulting distribution over actions (Eq.~\ref{eq_listener_action_rule}). 

\noindent\textbf{Evaluation Metrics.} We use four metrics to summarize speaker behavior. First, their probability of choosing a true utterance, P(truthful). Second, the probability of the listener choosing the optimal action in $A$ (Eq.~\ref{eq_listener_action_rule}), which we write as $\pi_L(a^*)$ for brevity. Third, the expected reward on the action context $A$ (Eq.~\ref{eq_combined_speaker}), which we write as $R_S(A)$. Finally, we want to know whether speakers are over-optimizing for a particular context. To evaluate this, we calculate the \textit{expected generalization}, which indicates whether the listener's resulting belief state yields good performance on other contexts drawn from $\mathcal{A}$. Formally, we compute the expected reward of the speaker's utterance $u$ across all possible contexts:
\begin{equation}
R_S(u, w, \mathcal{A}) = \sum_{A \in [\mathcal{A}]^3} R_S(u, w, A) P(A)
\end{equation}
where $P(A)$ is the probability of an action context $A$; here, we assume a uniform distribution over all contexts of size 3. Again, we shorten this to $R_S(\mathcal{A})$ for clarity. If local performance substantially exceeds generalization, $R_S(A) \gg R_S(\mathcal{A})$, we say the speaker generalizes poorly: it is optimizing local decision-making by providing false or less broadly useful information. 

\subsection{Simulation 1: Reproducing reference games}

\begin{figure}[t!]
\includegraphics[width=8.5cm]{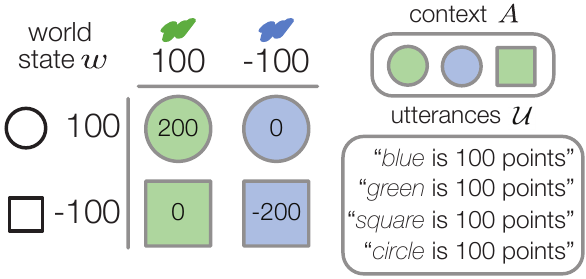}
\vspace{-1em}
\caption{Simulation 1 constructs a traditional Lewis signaling game within the signaling bandits framework. Note that two of the possible utterances are false.} 
\label{fig_reference_game}
\end{figure} 

\begin{table}[t!]
\centering
\begin{tabular}{l|llll}
Speaker & $P_S$(truthful)   & $\pi_L(a^*)$        & $R_S(A)$ & $R_S(\mathcal{A})$   \\ \hline
Belief  & 1.00  & .500    & 100   & - \\
Action  & 1.00  & .500    & 100   & - \\ 
Combined & 1.00 & .500    & 100   & -         
\end{tabular}
\caption{Simulation 1 results. Reference games align all three speakers' objectives and so cannot disambiguate them. Generalization, $R_S(\mathcal{A})$, does not apply to this setting.}
\label{tab:table_reference_game}
\end{table}

\label{lewis_signaling_section}
Our first simulation constructs a Lewis signaling game as a special case of our more general class of signaling bandits (Fig.~\ref{fig_reference_game}).
We show that this case cannot distinguish between our models, motivating Simulations 2 and 3.

\begin{figure*}[t!]
\begin{center}
\centering
\includegraphics[width=0.9\textwidth]{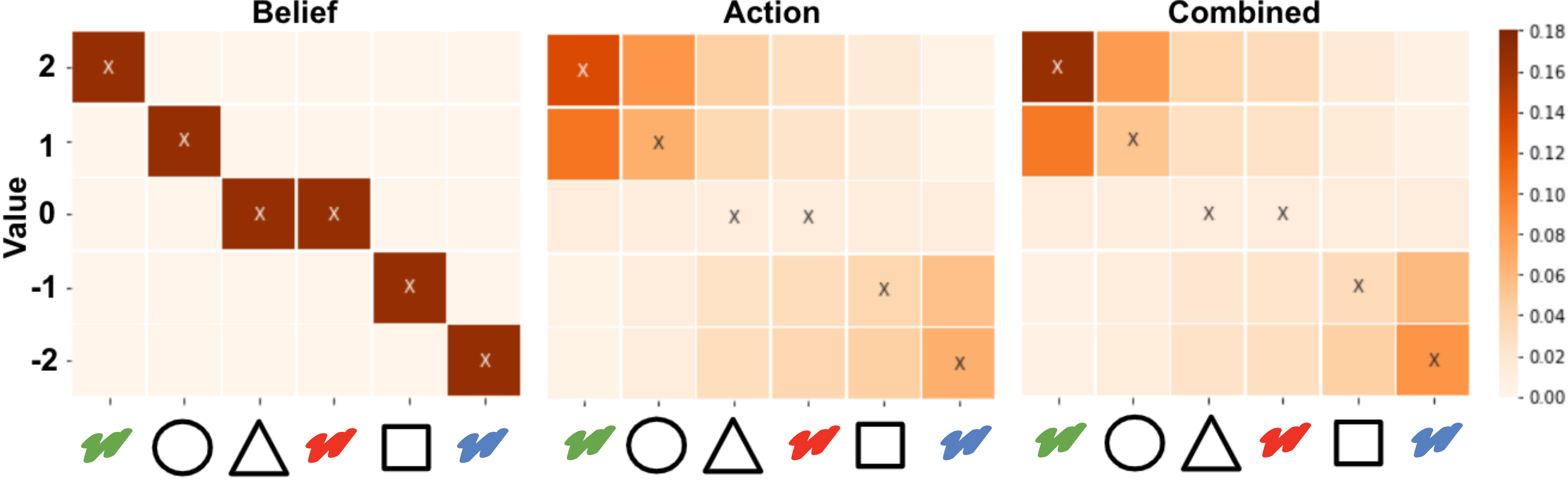}
\caption{Simulation 2 results. Shading indicates probability of speaker choosing that utterance, averaged over all 84 3-action contexts. X's indicate true utterances. Left: Belief speakers choose true utterances at random. Center/Right: Action and Combined speakers focus on decision-relevant features and exaggerate to improve listener decisions.}
\label{fig_myopic_utterance_choices}
\end{center}
\end{figure*}

\begin{table}[t!]
\centering
\begin{tabular}{l|llll}
Speaker & $P_S$(truthful) & $\pi_L(a^*)$ & $R_S(A)$ & $R_S(\mathcal{A})$ \\ \hline
Belief  & \textbf{1.00}        & .499              & .539 & \textbf{.539}      \\
Action  & .330        & \textbf{.772}              & 1.18   & .486    \\
Combined & .360        & .742              & \textbf{1.28} & .522   
\end{tabular}
\caption{Simulation 2 results. Action and Combined  frequently send false messages. They obtain high performance on the local context, $R_S(A)$, but generalize poorly, $R_S(\mathcal{A})$.}
\label{tab:table_myopic_bandits}
\end{table}

\noindent\textbf{Setup.} 
To construct a Lewis game with a single target, we define $w$ and construct a context $A$ containing one action with arbitrarily high reward (the target) and two with zero reward (the distractors).
We restrict utterances $\mathcal{U}$ to positive messages (corresponding to possible referential labels).

\noindent\textbf{Results and Discussion.} Results are summarized in Table~\ref{tab:table_reference_game}. All three speaker objectives are aligned, and so we find that they behave identically: they choose between the two literally true messages $\langle$Green, 100$\rangle$ or $	\langle$Circle, 100$\rangle$. No model has any reason to prefer a false utterance, or to prefer one true utterance over the other.

\subsection{Simulation 2: Divergent speaker behaviors}
\label{contextual_bandits_section}
We next explore how different speaker models may diverge for other tasks in our signaling bandit paradigm (Fig.~\ref{fig_game_features}). 

\noindent\textbf{Setup.} We consider the world state depicted in Fig.~\ref{fig_game_features}. We set $\mathcal{U}$ to all feature-value tuples and evaluate each speaker's behavior across all $\binom{9}{3}=84$ possible contexts of 3 actions. 

\begin{figure}[t!]
\centering
\includegraphics[width=7.9cm]{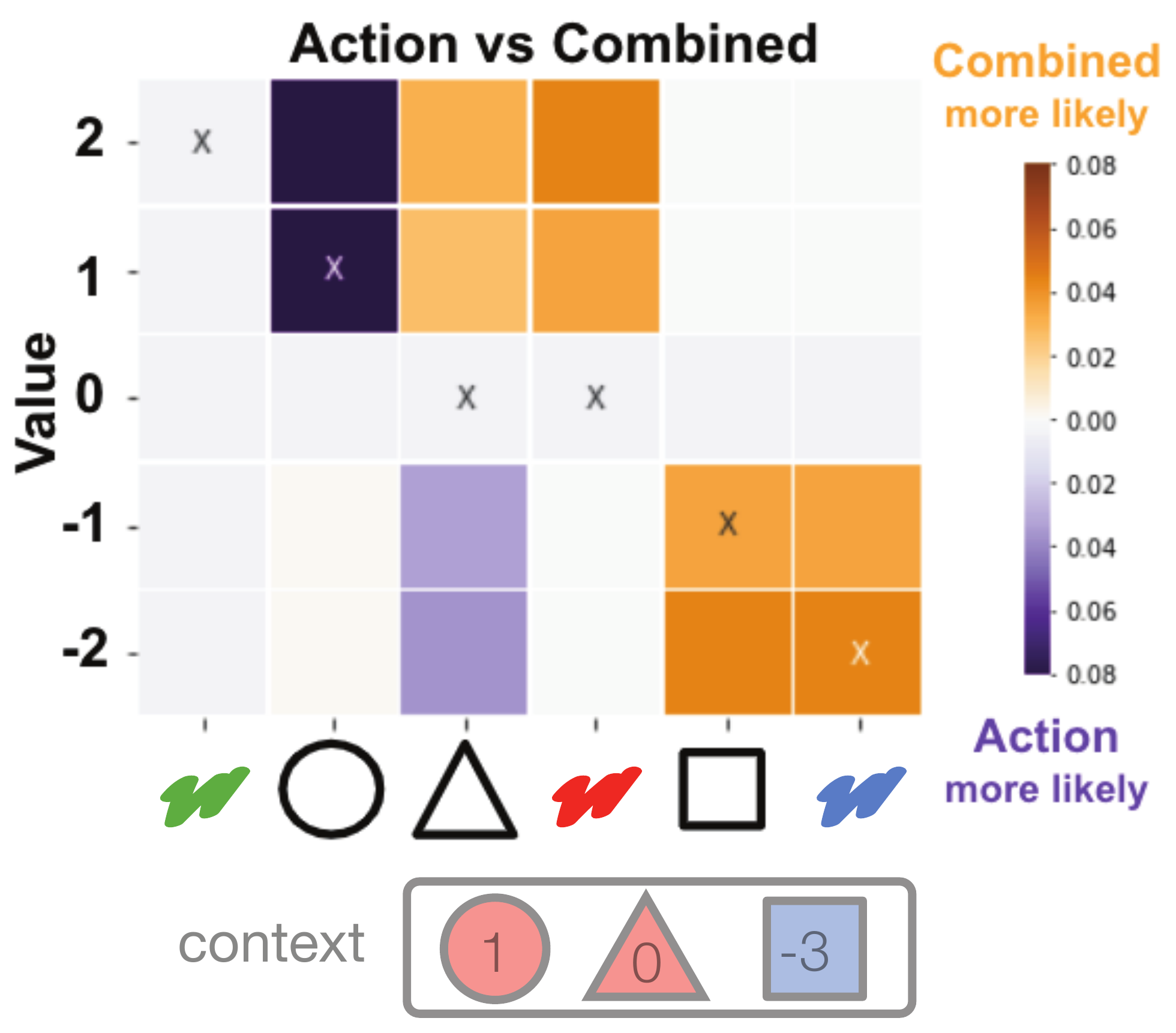}
\caption{Comparing Action and Combined in one action context (Simulation 2). Shading indicates difference in utterance probabilities between speakers. The Action speaker maximizes the listener's probability of choosing the red circle: $\pi_L(a^*)=.737, R_S(A)=.456$. The Combined speaker achieves higher reward by avoiding the blue square: $\pi_L(a^*)=.627, R_S(A)=.482$. Neither sends messages about green, demonstrating relevance effects.} 
\vspace{-2em}
\label{fig_action_vs_combined}
\end{figure}

\noindent\textbf{Results and Discussion.} We plot each speaker's probability over individual utterances in Fig.~\ref{fig_myopic_utterance_choices}, and summarize the results in Table~\ref{tab:table_myopic_bandits}. All metrics are averaged across the 84 contexts. 
First, we observe that Belief speakers choose a true utterance at random, regardless of the action context. This yields relatively poor performance locally but perfect generalization ($R_S(A)=R_S(\mathcal{A})=.539$). In contrast, both Action and Combined speakers lie frequently ($P_S$(truthful)$<.5$). Action speakers tailor their utterances to induce the single optimal action in each set. As a result, they exaggerate whichever feature values align  with the best action in the immediate context. This strategy succeeds locally: they induce the most-optimal action a majority of the time ($\pi_L(a^*)=.772$) and obtain more than twice the reward obtained by the Belief speaker ($R_S(A)=1.18$). However, the resulting distortion in listener beliefs means they generalize poorly ($R_S(\mathcal{A})=.486$). Finally, Combined speakers achieve a middle ground. They obtain the best outcome less frequently than Action speakers ($\pi_L(a^*)=.742$), but higher reward locally ($R_S(A)=1.28$) and generally ($R_S(\mathcal{A})=.522$). 
Sensitivity to the rewards of all three actions leads them to distort beliefs less than Action speakers. We visualize the divergence between Action and Combined speakers in Fig.~\ref{fig_action_vs_combined} in a single action context to better understand these differences.

\begin{table}[t!]
\centering
\begin{tabular}{l|llll}
Speaker & $P_S$(truthful)       & $\pi_L(a^*$)    &$R_S(A)$   & $R_S(\mathcal{A})$  \\ \hline
Belief  & \textbf{1.00}     & .499          & -         & .539       \\
Action  & .440          & .566          & -         & .748      \\
Combined & .534              & \textbf{.627} & -         & \textbf{.949}
\end{tabular}
\caption{Simulation 3 results. When communicating about a larger action context, Action and Combined speakers become more truthful and generalization improves dramatically.}
\label{tab:table_wise_bandits}
\end{table}
\tabcolsep=1.4mm

\subsection{Simulation 3: Expanding speaker context}
\label{meta_contextual_bandits_section}
Simulation 2 showed that Action and Combined speakers can be myopic: they produce messages to induce locally-optimal actions at the cost of generalization. Simulation 3 explores how this changes when they optimize over the entire action space $\mathcal{A}$. We find that both Action and Combined speakers become more truthful and generalize better.

\noindent\textbf{Setup.} We use the same world as Simulation 2 (shown in Fig.~\ref{fig_game_features}). We first construct a single ``global'' action context of all 9 actions: $A=\mathcal{A}$. We compute each speaker's distribution over utterances for this 9-action context, then evaluate generalization over 3-action contexts, $R_S(\mathcal{A})$. Because there is no ``local'' context, we do not compute $R_S(A)$. 

\noindent\textbf{Results and Discussion.} Results are summarized in Table~\ref{tab:table_wise_bandits}. The Belief speaker is unchanged. Both action-oriented speakers are more truthful and generalize better than Simulation 2. However, the Action speaker fixates on the \textit{single best} option (the green circle). It sends false messages which exaggerate its value, e.g. $\langle$Circle, 2$\rangle$, or discourage alternatives, e.g. $\langle$Red, -2$\rangle$. As a result, it fares poorly whenever a green circle is not present. This illustrates the brittleness of optimizing to obtain a specific action. Because the Combined speaker optimizes in expectation over all actions, it is more likely to send true messages about extreme values, e.g. $\langle$Green, 2$\rangle$ or $\langle$Blue, -2$\rangle$. It obtains both higher rewards $R_S(\cal{A})$ and the optimal action $\pi_L(a^*)$ more frequently.

\subsection{Effects of speaker optimality}

While we fixed the speaker optimality parameter $\beta_S$ throughout our simulations, it may interact in important ways with our model comparison.  
First, our Belief speaker is insensitive to $\beta$: since we assumed that the listener's prior over $w$ is uniform, all true utterances are equally valuable. 
By contrast, the Action and Combined speakers are sensitive to $\beta$ in different ways, since they take the soft-max over different quantities (Action over log-probabilities and Combined over expected utilities). 
At any given $\beta$, Combined dominated Action on all metrics, so we tuned $\beta$ to equate them ($\beta_\text{Action} = 3$, $\beta_\text{Combined}=2$). 
At the same time, it is informative to explore their asymptotic behavior as $\beta \rightarrow \infty$, which is summarized in Table~\ref{tab:table_optimal_bandits}. In Simulation 2, we find that Action and Combined still send false messages and generalize poorly in the limit. Yet when multiple messages are equally likely to induce an optimal action, the Combined speaker converges to preferring truthful ones while the Action speaker is indifferent. This difference results in dramatically better performance in Simulation 3: the Combined speaker converges on producing the utterance $\langle$Green, 2$\rangle$, while the optimal Action speaker remains ambivalent between $\langle$Green, 2$\rangle$ and $\langle$Circle, 2$\rangle$. In sum, the Combined speaker's sensitivity to underlying reward structure leads it to consistently produce more truthful and generalizable utterances across contexts and in the limit of optimality.

\begin{table}[t]
\centering
\begin{tabular}{l|l|llll}
Context & Speaker & $P_S$(truthful) & $\pi_L(a^*$)  & $R_S(A)$ & $R_S(\mathcal{A})$ \\ \hline
Both & Belief  & \textbf{1.00}  & .499          & .539      & .539 \\\hdashline
Local & Action  &  .458         & \textbf{.942}          & 1.63      & .690    \\
Local & Combined &  .488        & \textbf{.942}          & \textbf{1.67}      & .710   \\\hdashline
Global & Action  &  .500        & .624          & -         & .958    \\
Global & Combined &  \textbf{1.00}   & .731     & -         & \textbf{1.28}  
\end{tabular}
\caption{Speaker behavior as $\beta \rightarrow \infty$. \emph{Local} context corresponds to Simulation 2, and \emph{Global} context to Simulation 3.}
\label{tab:table_optimal_bandits}
\end{table}

\section{General Discussion}
Humans communicate to influence one anothers' beliefs \emph{and} actions. Here, we explored different ways speakers can reason about these objectives. We introduced two action-oriented speaker models which optimize for a downstream \emph{decision problem}, grounding relevance in the listener's actual decision context~\cite{roberts2012information}. Critically, we proposed that rational ``Combined'' speakers should consider both beliefs and actions: they should communicate to induce belief states that are likely to produce high-value actions. To distinguish speaker models, we introduced a new communication game, signaling bandits. Signaling bandits generalizes Lewis signaling games to multi-armed bandits, formalizing communication in richer decision settings.
Simulations show that the Combined speaker prefers \emph{generalizable} information that is likely to produce high-value actions across a distribution of possible future contexts.
This finding raises intriguing connections to belief-oriented accounts of generics \cite<e.g.``Birds fly'';>{tessler2019language} as well as biases towards generalizable examples in non-linguistic pedagogy~\cite{csibra2009natural, tomasello2016cultural}. 
We are thus optimistic that such speaker models may provide a bridge from communicative principles to social learning more broadly.

This work represents a small step towards a deeper exploration of action-grounded models of rational communication. First, human experiments are needed to validate our simulations. Second, we considered only literal listeners in collaborative settings. Pragmatic listeners may reason about a speaker's objectives and knowledge, as well as the action context the speaker considered \cite{goodman2013knowledge}. Finally, we explored only single-round gameplay; iterated games would allow for richer interactions. Speakers could observe listener actions and infer their beliefs via inverse reinforcement learning~\cite{ng2000algorithms}. A single message followed by learner actions in multiple contexts would force speakers to optimize for a distribution over contexts. This would make generalization an objective rather than an incidental effect, as in optimal reward design~\cite{singh2009rewards}. Listeners could learn both socially (via speaker messages) and individually (via their own actions). We hope we have successfully signaled the high value of research in this paradigm!


\section{Acknowledgements}
We thank our anonymous reviewers for their thoughtful feedback. This work was supported by NSF grant \#1545126 and John Templeton Foundation grant \#61454 to TLG and NSF grant \#1911835 to RDH. 

\vspace{2em}
\fbox{\parbox[b][][c]{7.3cm}{\centering {Code for simulations available at: \\
\href{https://github.com/tsumers/signaling-bandits}{\url{https://github.com/tsumers/signaling-bandits}}
}}}
\vspace{2em} \noindent

\nocite{clark1996using}

\bibliographystyle{apacite}

\setlength{\bibleftmargin}{.125in}
\setlength{\bibindent}{-\bibleftmargin}

\bibliography{references}

\end{document}